\documentclass[runningheads]{llncs}
\usepackage{graphicx}
\usepackage{subcaption}
\usepackage{amsmath,amsfonts,amssymb}
\usepackage{breqn}
\usepackage{tabularx}
\usepackage{multirow}
\usepackage{color}
\usepackage{tcolorbox}
\usepackage{bm}
\usepackage{CJK}
\usepackage{adjustbox}
\usepackage{ctable} 
\tcbset{width=0.9\textwidth,boxrule=0pt,colback=red,arc=0pt,auto outer arc,left=0pt,right=0pt,boxsep=5pt}

\usepackage[noend]{algpseudocode}
\usepackage[misc]{ifsym}
\usepackage[linesnumbered,ruled,vlined]{algorithm2e}
\SetKwInput{KwInput}{Input}                
\SetKwInput{KwOutput}{Output}              

\DeclareMathSymbol{\mh}{\mathord}{operators}{`\-}

\begin{document}

%
\title{Continual Learning with Knowledge Transfer for Sentiment Classification}

%
%

\author{Zixuan Ke\inst{1} \and
Bing Liu\inst{1} \and
Hao Wang\inst{2}\and
Lei Shu\inst{1}}

%
%

\institute{
University of Illinois at Chicago, USA \\
\email{\{zke4,liub\}@uic.edu, shulindt@gmail.com}\\ \and
Southwest Jiaotong University, Chengdu, China\\
\email{cshaowang@gmail.com}}

\maketitle
\begin{abstract}
This paper studies continual learning (CL) for sentiment classification (SC). In this setting, the CL system learns a sequence of SC tasks incrementally in a neural network, where each task builds a classifier to classify the sentiment of reviews of a particular product category or domain. 
Two natural questions are: Can the system transfer the knowledge learned in the past from the previous tasks to the new task to help it learn a better model for the new task? And, can old models for previous tasks be improved in the process as well?  
This paper proposes a novel technique called KAN to achieve these objectives. KAN can markedly improve the SC accuracy of both the new task and the old tasks via forward and backward knowledge transfer. The effectiveness of KAN is demonstrated through extensive experiments\footnote{Code and data are available at: \url{https://github.com/ZixuanKe/LifelongSentClass}
}. 
\end{abstract}

\section{Introduction}
\label{sec:introduction}

Continual learning (CL) aims to learn a sequence of tasks incrementally~\cite{chen2018lifelong,Parisi2019continual}. Once a task is learned, its training data is typically forgotten. The focus of the existing CL research has been on solving the catastrophic forgetting (CF) problem~\cite{chen2018lifelong,Parisi2019continual}. 
CF means that when a neural network learns a sequence of tasks, the learning of each new task is likely to change the network weights or parameters learned for previous tasks, which degrades the model accuracy for the previous tasks~\cite{mccloskey1989catastrophic}. 
There are two main CL settings in the existing research: 

\textbf{Class continual learning (CCL)}: In CCL, each task consists of one or more classes to be learned. Only one model is built for all classes seen so far. In testing, a test instance from any class may be presented to the model for it to classify without giving it any task information used in training. 

\textbf{Task continual learning (TCL)}. In TCL, each task is a separate classification problem (e.g., one classifying different breeds of dogs and another classifying different types of birds). TCL builds a set of classification models (one per task) in one neural network. In testing, the system knows which task each test instance belongs to and uses only the model for the task to classify the test instance. 

In this paper, we work in the TCL setting to continually learn a sequence of sentiment analysis (SA) tasks. 
Typically, a SA company has to work for many clients and each client wants to study public opinions about one or more categories of its products/services and those of its competitors. The sentiment analysis of each category of products/services is a task. For confidentiality, a client often does not allow the SA company to share its data with or use its data for any other client. Continual learning is a natural fit. In this case, we also want to improve the SA accuracy over time without breaching confidentiality. This presents two key challenges: (1) how to transfer the knowledge learned from the previous tasks to the new task to help it learn better without using the previous tasks' data, and (2) how to improve old models for the previous tasks in the process without CF? In~\cite{DBLP:conf/dasfaa/LvWLCZ19}, the authors showed that CL can help improve the accuracy of document-level sentiment classification (SC), which is a sub-problem of SA~\cite{liu2015sentiment}. In this paper, we propose a significantly better model, called KAN (\textit{Knowledge Accessibility Network}). Note that each task here is a two-class SC problem, i.e., classifying whether a review for a product is positive or negative. 

A fair amount of work has been done on CL. However, existing techniques have mainly focused on dealing with catastrophic forgetting (CF)~\cite{chen2018lifelong,Parisi2019continual}. In learning a new task, they typically try to make the weights update toward less harmful directions to previous tasks, or to prevent the important weights for previous tasks from being significantly changed. We will detail these and other related work in the next section. 
Dealing with only CF is far from sufficient for SC. In most existing studies of CL, the tasks are quite different and have little shared knowledge. It thus makes sense to focus on dealing with CF. However, for SC, the tasks are similar because words and phrases used to express sentiments for different products/tasks are similar. As we will see in Section~\ref{sec:results}, CF is not a major problem in CL for SC due to the shared knowledge across tasks. Our main goal is thus to leverage the shared knowledge among tasks to perform significantly better than learning individual tasks separately in isolation. 


To achieve the goal of leveraging the shared knowledge among tasks to improve the SC accuracy, KAN uses two sub-networks, the \textit{main continual learning} (MCL) network and the \textit{accessibility} (AC) network. The core of MCL is a \textit{knowledge base} (KB), which stores the knowledge learned from all trained tasks. In learning each new task, the AC network decides which part of the past knowledge is useful to the new task and can be shared. This enables \textit{forward knowledge transfer}. 
Also importantly, the shared knowledge is enhanced during the new task training using its data, which results in \textit{backward knowledge transfer}. Thus, KAN not only improves the model accuracy of the new task but also improves the accuracy of the previous tasks without any additional operations. Extensive experiments show that KAN markedly outperforms state-of-the-art baselines. 

\section{Related Work}
\label{sec:related}

Continual learning (CL) has been researched fairly extensively in machine learning (see the surveys in~\cite{chen2018lifelong,Parisi2019continual}). Existing approaches have primarily focused on dealing with catastrophic forgetting (CF). Lifelong learning is also closely related~\cite{Silver2013,ruvolo2013ella,chen2014topic,chen2018lifelong}, which mainly aims to improve the new task learning through forward knowledge transfer. We discuss them in turn and also their applications in sentiment classification (SC). 

\vspace{2mm}
\noindent
\textbf{Continual Learning.} Several approaches have been proposed to deal with CF: 



\textit{Regularization-based methods,} such as those in~\cite{Kirkpatrick2017overcoming,DBLP:conf/nips/LeeKJHZ17,Seff2017continual}, add a regularization in the loss function to consolidate previous knowledge when learning a new task. 

\textit{Parameter isolation-based methods,} such as those in \cite{DBLP:conf/icml/SerraSMK18,DBLP:conf/cvpr/MallyaL18,fernando2017pathnet}, make different subsets of the model parameters dedicated to different tasks. They identify the parts of the network that are important to the previous tasks and mask them out during the training of the new task.

\textit{Gradient projection-based methods}, such as that in~\cite{zeng2019continuous}, ensure the gradient updates occur only in the orthogonal direction to the input of the previous tasks. Then, the {\color{black}weight updating} for the new task have little effect on the weights for the previous tasks. 

\textit{Exemplar and replay-based methods}, such as those in~\cite{Rebuffi2017,Lopez2017gradient,Chaudhry2019ICLR}, retain an exemplar set that best approximates the previous tasks to help train the new task. The methods in~\cite{Shin2017continual,Kamra2017deep,Rostami2019ijcai,He2018overcoming} instead took the approach of building data generators for the previous tasks so that in learning the new task, they can use some generated data for previous tasks to help avoid forgetting. 

As these methods are mainly for avoiding CF, after learning a sequence of tasks, their final models are typically worse than learning each task separately. The proposed KAN not only deals with CF, but also perform forward and backward transfer to improve the accuracy of both the past and the new tasks. 

To our knowledge, SRK~\cite{DBLP:conf/dasfaa/LvWLCZ19} is the only CL method for sentiment classification (SC). It consists of two networks, a feature learning network and a knowledge retention network, which are combined to perform CL. However, SRK only does forward transfer as it protects the past knowledge and thus cannot do backward transfer as KAN does. More importantly, due to this protection, its forward transfer also suffers because it cannot adapt the previous knowledge but only use it without change. Its results are thus poorer than KAN. The SRK paper also showed that CF is not a major issue for continual SC learning as the SC tasks are highly similar, which also explains why adaption of the previous knowledge in forward transfer in KAN does not cause CF. 


{\color{black}KAN is closely related to the TCL system HAT \cite{DBLP:conf/icml/SerraSMK18} as HAT also trains a binary mask using hard attention. However, HAT's hard attention is for identifying what past knowledge in the network should be protected for each previous task so that the new task learning will not modify this previous knowledge. This is effective for avoiding CF, not appropriate for our SC tasks due to the shared knowledge across tasks in SC. KAN trains an accessibility mask for the current/new task to decide what previous knowledge can be accessed by or shared with the current task to enable both forward and backward knowledge transfer. There is no concept of knowledge transfer in HAT. In terms of architecture, KAN has two sub-networks: the accessibility (AC) network and the main continual learning (MCL) network, while HAT has only one - it does not have the AC network. KAN's AC network trains the AC mask to determine what knowledge can be shared. The MCL network stores the knowledge and applies the trained AC mask to the knowledge base. This setting enables KAN not only to adapt the shared  knowledge across tasks to produce more accurate models, but also to avoid CF. HAT has only one network, and its mask is to block only the knowledge that is important to previous task models. 
} 

\vspace{2mm}
\noindent
\textbf{Lifelong Learning (LL) for SC.} 
The authors of~\cite{DBLP:conf/acl/ChenM015,hao2019forward} proposed a Naive Bayes (NB) approach to help improve the new task learning. A heuristic NB method was also proposed in~\cite{hao2019forward}.  
\cite{xia2017distantly} presented {\color{black}a} LL approach based on voting of individual task classifiers. 
All these works do not use neural networks, and are not concerned with the CF problem. The work in \cite{ShuXuLiu2017,shuai2018lifelong} uses LL for aspect-based sentiment analysis, which is an entirely different problem than document-level sentiment classification (SC) studied in this paper. 

\section{Proposed Model KAN}
\label{sec:model}


To improve the classification accuracy and also to avoid forgetting, we need to identify some past knowledge that is shareable and update-able in learning the new task so that no forgetting of the past knowledge will occur and both the new task and the past tasks can improve. 

We solve this problem by taking inspiration from what we humans seem to do. For example, we may ``forget'' our phone number 10 years ago, but if the same number or a similar number shows up again, our brain may quickly retrieve the old phone number and make both the new and old numbers memorized more strongly. Biological research \cite{kornell2009unsuccessful} has shown that our brain keeps track of the knowledge accessibility. If some parts of our previous knowledge are useful to the new task (i.e., shared knowledge between the new task and some previous tasks), our brain sets those parts accessible to enable {\em forward knowledge transfer}. This also enables {\em backward knowledge transfer} as they are now accessible and we have the opportunity to strengthen them based on the new task data. For those not useful parts of the previous knowledge, they are set to inaccessible, which protects them from being changed. Inspired by this idea, we design a memory and accessibility mechanism. 

\begin{figure}[t]
\centering
\includegraphics[width=0.8\columnwidth]{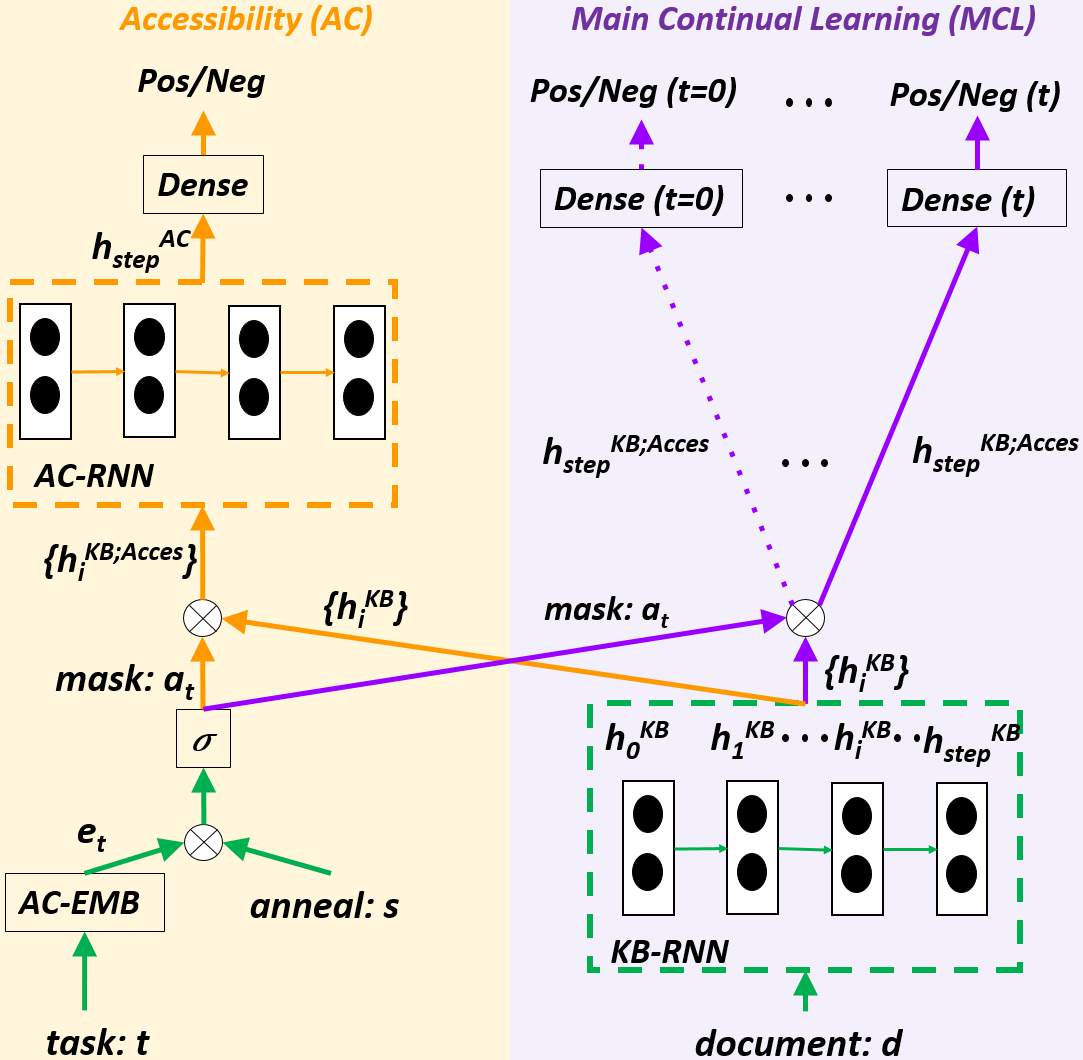} 
\vspace{-3mm}
\caption{The KAN architecture (best viewed in color). The purple box contains the main continual learning (MCL) component, and the yellow box contains the accessibility (AC) component. The green arrows represent forward paths shared by both components. The yellow and purple arrows represent the forward paths used only in AC and MCL respectively.}
\label{overview}
\vspace{-3mm}
\end{figure}

{\color{black}We need to solve two key problems}: (1) how to detect the accessibility of the knowledge in the memory (which we call \textit{knowledge base} (KB)), i.e., identifying the part of the previous knowledge that is useful to the new task; (2) how to leverage the identified useful/shared knowledge to help the new task learning while also protecting the other part. To address these challenges, we propose the \textit{Knowledge and Accessibility Network} (KAN) shown in Figure~\ref{overview}.

\subsection{Overview of KAN}

KAN has two components (see Figure~\ref{overview}), 
the {\em main continual learning} (MCL) component in the purple box and the {\em accessibility} (AC) component in the yellow box. MCL performs the main continual learning and testing (AC is not used in testing except the mask $\bm{a}_t$ generated from the task id $t$). We see the sentiment classification heads (pos/neg) for each task at the top. Below them are the dense layers and further down is the \textit{knowledge base} (KB) (the memory) in the green dash-lined box. KB, which is modeled using an RNN (we use GRU in our system) and is called KB-RNN, contains both the task-specific and shared knowledge across tasks. AC decides which part of the knowledge (or units) in the KB is accessible by the current task $t$ by setting a {\color{black}\textit{binary task-based mask}} $\bm{a}_t$. Each task is indicated by its task embedding produced by AC-EMB from the task id ($t$). AC-EMB is a randomly initialized embedding layer. The inputs to KAN are the task id $t$ and document $d$. They are used in training both components via the mask $\bm{a}_t$ and $\{\bm{h}_i^{KB}\}$ (hidden states in KB-RNN) links. 

\vspace{+4mm}
\noindent
\textbf{AC Training.} In the AC training phase, only Task Embedding (AC-EMB), AC-RNN and others in the yellow box are trainable. Consider the KB has already retained the knowledge learned from tasks $0...t \mh 1$. When the new task $t$ arrives, we first train the AC component to produce a binary task-based mask $\bm{a}_t$ (a matrix with the same size as the KB $\{\bm{h}_i^{KB}\}$) to indicate which units in KB-RNN are accessible for the new/current task $t$. Since the mask is trained based on the new task data with the previous knowledge in KB-RNN fixed, those KB-RNN units that are not masked (meaning that they are useful to the new task) are the accessible units with their entries in $\bm{a}_t$ as 1 (unmasked). The other units are inaccessible with their entries in $\bm{a}_t$ as 0 (masked). 

\vspace{+4mm}
\noindent
\textbf{MCL Training.} After AC training, we start MCL training. In this phase, only KB-RNN and others in the purple box are trainable. The trained binary task-based mask $\bm{a}_t$ is element-wise multiplied by the output vectors of KB-RNN. This operation protects those inaccessible units since no gradient flows across them while allowing those accessible units to be updated since the mask does not stop gradients for them. This clearly enables \textit{forward knowledge transfer} because those accessible units selected by the mask represent the knowledge from the previous tasks that can be leveraged by the current/new task.  It also enables {\color{black}\textit{backward knowledge transfer} and avoidance of CF} because (1) if the accessible units are not important to the previous tasks, any modification to them does not degrade the previous tasks' performance, and (2) if the accessible units are useful for some previous tasks, updating them enable them to improve as we now have more data to enhance the shared knowledge. 

\begin{algorithm}[H]
\DontPrintSemicolon
  \KwInput{Dataset $D = (D_0, ..., D_T)$}
  \KwOutput{Parameters of KB-RNN $\bm{W}^{KB \mh RNN}$, of AC-RNN $\bm{W}^{AC \mh RNN}$, and of AC-EMB $\bm{W}^{AC \mh EMB}$}
  \For{$t = 0,..,T $} 
  { 
      \If{$t = 0$} 
        {
           $\bm{W}^{KB \mh RNN}_0 = MCLtraining(D_0)$  \label{TrainingAN:line:start:t=1}

           $\bm{W}^{AC \mh RNN}_0,\bm{W}^{AC \mh EMB}_0$ = $ACtraining(D_0,\bm{W}_0^{KB \mh RNN})$ \label{TrainingAN:line:end:t=1}
        } 

    \Else 
    {
           $\bm{W}^{AC \mh RNN}_t,\bm{W}^{AC \mh EMB}_t$ = $ACtraining(D_t,\bm{W}_{t \mh 1}^{KB \mh RNN})$ \label{TrainingAN:line:start:t>1}
           
           $\bm{W}^{KB \mh RNN}_t$ = $MCLtraining(D_t,\bm{W}^{AC \mh EMB}_t)$ \label{TrainingAN:line:end:t>1}
    }  	
  }  
\caption{Continual Learning in KAN}
\label{TrainingAN}
\end{algorithm}

\vspace{+4mm}
\noindent
\textbf{Continual Learning in KAN:}  The algorithm for continual learning in KAN is given in Algorithm~\ref{TrainingAN}. For each new task, AC training is done first and then MCL training.
An exception is at the first task (lines \ref{TrainingAN:line:start:t=1}-\ref{TrainingAN:line:end:t=1} in Algorithm~\ref{TrainingAN}). At the very beginning, KB has no knowledge, and thus nothing can be used by AC. Therefore, we train MCL (and KB) before AC to obtain some knowledge first. However, after the first task, AC is always trained before MCL (and KB) for each new task (lines \ref{TrainingAN:line:start:t>1}-\ref{TrainingAN:line:end:t>1}).




\subsection{Details of Accessibility Training}
\label{sec:AC}

AC training aims to detect the accessibility of the knowledge retained in the KB given the new task data. As shown in Figure~\ref{overview} and Algorithm~\ref{TrainingAC} \footnote{For simplicity, we only show the process for one training example, but our actual system trains in batches.}, it takes as inputs a training example $d$ and the trained knowledge base KB-RNN $\bm{W}^{KB \mh RNN}_{t-1}$, which has been trained because AC is always trained after KB (KB-RNN) was trained on the previous task.
To generate a binary matrix $\bm{a}_t$ so that it can work as a mask in both the MCL (and KB) and AC training phases, we borrow the idea of {\em hard attention} in \cite{DBLP:conf/icml/SerraSMK18,DBLP:conf/icml/XuBKCCSZB15,aharoni2017} where the values in the attention matrix are {\em binary} instead of a probability distribution as in {\em soft attention}.

\vspace{+4mm}
\noindent
\textbf{Hard Attention Training.} Since we can access the task id in both training and testing, a natural choice is to leverage the task embedding to compute the hard attention. Specifically, for a task id $t$, we first compute a task embedding $\bm{e}_t$ by feeding the task id into the task embedding layer (AC-EMB) where a 1-hot {\em id} vector is multiplied by a {\em randomly initialized} parameter matrix. Using the resulting task embedding $\bm{e}_t$, we apply a gate function $\sigma(x) \in [0,1]$ and a positive scaling parameter $s$ to compute the binary attention $\bm{a}_t$ as shown in lines \ref{TrainingAC:start:a_t} and \ref{TrainingAC:end:a_t} in Algorithm~\ref{TrainingAC}. Intuitively, it uses a unit step function as the activation function $\sigma(x)$. However, this is not ideal because it is non-differentiable. We want to train the embedding $\bm{e}_t$ with back propagation. Motivated by \cite{DBLP:conf/icml/SerraSMK18}, we use a {\em sigmoid} function with a positive scaling parameter $s$ to construct a pseudo-step function allowing the gradient to flow. This scaling parameter is introduced to control the polarization of our pseudo-step function and the output $\bm{a}_t$. Our strategy is to anneal $s$ during training, inducing a gradient flow and set $s=s_{max}$ during testing. This is because using a {\color{black}hyperparameter} $s_{max} \gg 1$, we can make our sigmoid function approximate to a unit step function. Meanwhile, when $s \to \infty$ we get $\bm{a}_t \to \{0,1\}$ and when $s \to 0$, we get $\bm{a}_t \to 0.5$. We start the training epoch with all units being equally active by using the latter and progressively polarize them within the epoch. Specifically, we anneal $s$ as follows:
\begin{dmath}
s = \frac{1}{s_{max}} + (s_{max} - \frac{1}{s_{max}})\frac{b-1}{B-1}
\label{eq:anneals}
\end{dmath}
where {\color{black}$b = 1,...B$} is the batch index and $B$ is the total number of batches in an epoch.

To better understand the hard attention training, recall that the resulting mask $\bm{a}_t$ needs to be binary so that it can be used to block/unblock some units' training in both phases. The task id is used to control the mask to condition the KB. To achieve this, we need to make sure the embedding of the task id is trainable for which we adopt sigmoid as the pseudo gate function. The training procedure is annealing: at first, {\color{black}$s\to 0$ ($s=\frac{1}{s_{max}}$)} and the mask is still a soft attention. After certain batches, $s$ becomes large and $\sigma(s \otimes \bm{e}_t)$  becomes very similar to a gate function. After training, those units with their entries in $\bm{a}_t$ as 1 are accessible to task $t$ while the others are inaccessible. Another advantage of training a hard attention is that we can easily retrieve the binary task-based $\bm{a}_t$ after training: we simply adopt $s_{max}$ to be $s$ and apply $\sigma(s_{max} \otimes \bm{e}_t)$.

\vspace{+4mm}
\noindent 
\textbf{Apply Hard Attention to the Network.} The fixed KB-RNN takes a training example $d$ as input and produces a sequence of output vectors $\{\bm{h}_i^{KB}\}$ which have incorporated the previous knowledge (line \ref{TrainingAC:EncodeExample} in Algorithm~\ref{TrainingAC}). $i$ denotes the $i$th step of the RNN or the representation of the $i$th word of the input, {\color{black}ranging from $0$ to the step size $step$}. $\{\bm{h}_i^{KB}\}$ then performs element-wise multiplication with the task-based mask $\bm{a}_t$ to get the accessibility representation of the previous knowledge $\{h_i^{KB;Access}\}$ (line \ref{TrainingAC:Element-wiseMultiplication}). Recall that $\bm{e}_t$ is a task id embedding vector and $\bm{e}_t \in \mathbb{R}^{dim}$, where $dim$ refers to the dimension size, {\color{black}and therefore $\bm{a}_t \in \mathbb{R}^{dim}$ ($\bm{a}_t = \sigma(s \otimes \bm{e}_t)$)}. In other words, we expand the vector $\bm{a}_t$ (repeat the vector $step$ times) to match the {\color{black}size} of $\{\bm{h}_i^{KB}\}$, and then perform element-wise multiplication. This accessibility representation encodes the useful knowledge from the previous tasks. We first feed the representation into AC-RNN to learn some additional new task knowledge (line \ref{TrainingAC:EncodeAccessible}). The last step of the resulting sequence of vectors $\{\bm{h}_i^{AC}\}$, which is {\color{black}$\bm{h}_{step}^{AC}$,} then goes through a dense layer to reduce the vector's dimension and finally compute the loss based on cross entropy (line \ref{TrainingAC:classification}).

\begin{algorithm}[H]
\DontPrintSemicolon
  \KwInput{A training example $d$, scaling parameter $s$, task id $t$, trained KB-RNN $\bm{W}^{KB \mh RNN}_{t \mh 1}$}
  \KwOutput{Parameters of AC-RNN $\bm{W}^{AC \mh RNN}_t$ and of AC-EMB $\bm{W}^{AC \mh EMB}_t$}

  \setcounter{AlgoLine}{0}

  $\bm{e}_t = $AC-EMB$(t)$  \label{TrainingAC:start:a_t}   \tcp*[h]{{\color{black}AC-EMB is trainable.}}

  $\bm{a}_t = \sigma(s \otimes \bm{e}_t)$  \label{TrainingAC:end:a_t}   
  \tcp*[h]{{\color{black}We anneal $s$ as shown in Eq.\ref{eq:anneals}}}

  $\{\bm{h}^{KB}_i\} = $KB-RNN$(d)$  \label{TrainingAC:EncodeExample}   
  \tcp*[h]{{\color{black}KB-RNN is already trained and is fixed.}}
  
  \{$\bm{h}^{KB;Access}_i\} = \{\bm{h}^{KB}_i\} \otimes \bm{a}_t$  \label{TrainingAC:Element-wiseMultiplication} 

  \{$\bm{h}^{AC}_i\} = $AC-RNN(\{$\bm{h}^{KB;Access}_i$\}) \label{TrainingAC:EncodeAccessible} 
  \tcp*[h]{{\color{black}AC-RNN is trainable.}}
 
  $\mathcal{L}^{AC} = CrossEntropy(d_{label}, Dense(\bm{h}_{step}^{AC}))$ \label{TrainingAC:classification} 
  \tcp*[h]{{\color{black}Compute the AC loss.}}

\caption{AC Training}
\label{TrainingAC}
\end{algorithm}

\subsection{Details of Main Continual Learning Training}
\label{sec:KB}

MCL training learns the current task knowledge and protects the knowledge learned in previous tasks in the KB (KB-RNN). As shown in Algorithm~\ref{TrainingKB} and Figure~\ref{overview}, it takes an input training example $d$ in the corresponding dataset $D_t$ and encodes $d$ via KB-RNN (line \ref{TrainingKB:EncodeExample}), which results in a sequence of vectors \{$\bm{h}_i^{KB}$\}. Following our training scheme, i.e., AC-RNN and AC-EMB are always trained before KB for a new task $t$ (except for the first one), we already have the trained AC-EMB $\bm{W}^{AC \mh EMB}_t$ when discussing KB. We therefore can compute the mask $\bm{a}_t$ from $\bm{W}^{AC \mh EMB}_t$ (lines \ref{TrainingKB:start:a_t}-\ref{TrainingKB:end:a_t}). Note that {\color{black}we expand $\bm{a}_t$ to be} a binary matrix and $\bm{a}_t \in  \mathbb{R}^{step \times dim}$ where $step$ refers to step size of KB-RNN and $dim$ refers to the dimension size of the task embedding vector. For the first task (i.e., $t=0$), we simply assume $a_0$ as a matrix of ones (line \ref{TrainingKB:Seta_tOnes}).

\vspace{+4mm}
\noindent
\textbf{Block the Inaccessible and Unblock the Accessible Units.} Naturally, we want to ``remove'' those inaccessible units so that only accessible ones can contribute to the training of KB-RNN. An efficient method is to simply element-wise multiply the outputs of KB \{$\bm{h}_i^{KB}\}$ by $\bm{a}_t$ (line \ref{TrainingKB:Element-wiseMultiplication}). Since $\bm{a}_t$ is a binary mask, only those KB with mask 1 can be updated by backward propagation. This is equivalent to modifying the gradient $\bm{g}$ with the mask $\bm{a}_t$:

\begin{dmath}
\bm{g'} = \bm{a}_t \otimes \bm{g}
\end{dmath}

The resulting vectors can be seen as the representation of {\em accessible} knowledge \{$\bm{h}_i^{KB;Access}$\}. 
Finally, we take the last step of the accessible knowledge vectors {\color{black}\{$\bm{h}_i^{KB;Access}$\}, which is $\bm{h}_{step}^{KB;Access}$,} to a fully connected layer to perform classification (line \ref{TrainingKB:Classification}). Note that we employ {\em multi-head} configuration (upper components in the KB training phase in Figure~\ref{overview}) which means each task is allocated an exclusive dense layer. These dense layers are mapped to the dimension of the number of classes $c$ ($c$=2) so that we can use different dense layer to perform classification according to different task id. 

\begin{algorithm}[H]
\DontPrintSemicolon
  \KwInput{A training example $d$, task id $t$, and trained task embedding $\bm{W}_t^{AC \mh EMB}$}
  \KwOutput{Parameters of KB-RNN $\bm{W}^{KB \mh RNN}_t$}
  \setcounter{AlgoLine}{0}
  
  \{$\bm{h}^{KB}_i\} = $KB-RNN$(d)$  \label{TrainingKB:EncodeExample}
  \tcp*[h]{{\color{black}KB-RNN is trainable.}}

  \If{$t$ = 0}
    {Set all values of $\bm{a}_0$ to 1 \label{TrainingKB:Seta_tOnes}}
  \Else
    {
      $\bm{e}_t = $AC-EMB$(t)$  \label{TrainingKB:start:a_t}
      \tcp*[h]{{\color{black}AC-EMB is trainable.}}
      
      $\bm{a}_t = \sigma(s_{max} \otimes \bm{e}_t)$  \label{TrainingKB:end:a_t}   
      \tcp*[h]{{\color{black}Use $s_{max}$ to retrieve the trained mask.}}
    } 
    
  \{$\bm{h}^{KB;Acces}_i\} = \{\bm{h}^{KB}_i\} \otimes \bm{a}_t$  \label{TrainingKB:Element-wiseMultiplication}  

  $\mathcal{L}^{KB} = CrossEntropy(d_{label}, Dense(\bm{h}_{step}^{KB;Acces}))$ \label{TrainingKB:Classification}  
  \tcp*[h]{{\color{black}Compute the {\color{black}MCL} loss.}}

\caption{MCL Training}
\label{TrainingKB}
\end{algorithm}

\subsection{Comparing Accessibility and Importance}

Many existing models discussed in Section~\ref{sec:related} detect the {\em importance} of units. That is, they identify units or parameters that are important to previous tasks so that in learning the new task, the learner can protect them to avoid forgetting the previous tasks' models. However, it is also stifling the chance for adapting and updating the previous knowledge to help learn the new task in \textit{forward transfer} and for improving previous tasks to achieve {\em backward transfer}. In contrast, our concept of accessibility is very different, which is for the {\em current task}. KAN detects the accessibility of units and {\em safely} update the weights of the accessible units because of the shared knowledge. This enables adaptation in forward transfer. If some units are accessible for the current task, KAN can update them based on the {\em current task} training data. If those units are also accessible by some previous tasks, it suggests that there is some shared knowledge between the current and the previous tasks. This results in {\em backward transfer}. One can also see this as {\em strengthening the shared knowledge} since we now have more data for the shared knowledge training. In short, training the accessible units is helpful to both the current and the previous tasks.

\section{Experiments}
\label{sec:experiments}

We now evaluate KAN for continual document sentiment classification (SC). We follow the standard continual learning evaluation procedure~\cite{DBLP:journals/corr/abs-1909-08383} as follows. 
We provide a sequence of SC tasks with their training datasets for KAN to learn one by one. Each task learns to perform SC on reviews of a type of product. Once a task is learned, its training data is discarded. After all tasks are learned, we test all task models using their respective test data. In training each task, we use its validation set to decide when to stop training.   


\subsection{Experimental Data}

Our dataset consists of Amazon reviews from 24 different types of products, which make up the 24 tasks. Each task has 2500 positive (with 4 or 5 stars) and 2500 negative (with 1 or 2 stars) reviews. We further split the reviews in each task into training, testing and validation set in the ratio of 8:1:1. We didn't use the datasets in~\cite{DBLP:conf/dasfaa/LvWLCZ19} as they are all highly skewed with mostly positive examples. Without doing anything, the system can achieve more than $80\%$ of accuracy. We also did not use the commonly employed sentiment analysis datasets from SemEval~\cite{pontiki2016semeval} because its reviews involve only two types of products/services, i.e., laptop and restaurant, which are too few for continual learning. 

\subsection{Baselines}
\label{sec:baselines}
We consider a wide range of baselines: (1) isolated learning of each task; (2) state-of-the-art continual learning methods; (3) existing continual or lifelong sentiment classification models; and (4) a naive continual learning model.

\textbf{One task learning (ONE)} builds an isolated model for each task individually, independent of other tasks. There is no knowledge transfer. The network is the same as KAN but without AC and accessibility mask, consisting of word embeddings, a conventional GRU layer and a fully connected layer for classification. The same network is used in the other variant of KAN, i.e., N-CL. 

\textbf{Elastic Weight Consolidation (EWC)} \cite{Kirkpatrick2017overcoming} is a popular regularization-based continual learning method, which slows down learning for weights that are important to the previous tasks.

\textbf{Hard Attention to Task (HAT)} \cite{DBLP:conf/icml/SerraSMK18} learns pathways in a given base network using the given task id (and thus task incremental). The pathways are then used to obtain the task-specific networks. It is the state-of-the-art {\color{black}task continual learning (TCL) model} for avoiding catastrophic forgetting (CF).

\textbf{Orthogonal Weights Modification (OWM)} \cite{zeng2019continuous} is a state-of-the-art class continual learning (CCL) method. 
Since OWM is a CCL method but we need a TCL method, we adapt it for TCL. Specifically, we only train on the corresponding head of the specific task id during training and only consider the corresponding head's prediction during testing.  

\textbf{Sentiment Classification by Retained Knowledge (SRK)} \cite{DBLP:conf/dasfaa/LvWLCZ19} is the only prior work on continual sentiment classification. It achieves only limited forward transfer as we discussed {\color{black}in Section 2}. 

\textbf{Lifelong Learning for Sentiment Classification (LSC)}, proposed by \cite{DBLP:conf/acl/ChenM015}, is a Naive Bayes based lifelong learning approach to SC. It does forward transfer but not continual learning and thus has no CF issue. The main goal of this traditional method is to improve only the performance of the new task. 


\textbf{Naive continual learning (N-CL)} 
greedily trains a sequence of SC tasks incrementally without dealing with CF. Note that this is equivalent to KAN after removing AC and mask, i.e., it uses the same network as ONE.

\subsection{Network and Training}
Unless stated otherwise, we employ the embedding {\color{black} with 300 dimensions to represent the input text. GRU's with 300 dimensions are used for both AC-RNN and KB-RNN}. 
We adopt Glove 300d~\footnote{https://github.com/stanfordnlp/GloVe} as pre-trained word embeddings and fix them during training of KAN. The fully connected layer with softmax output is used as the final layer(s), together with categorical cross-entropy loss. During KAN training, we initialize the hidden state for each element in the batch to 0. We set $s_{max}$ to 140 in the $s$ annealing algorithm, dropout to 0.5 between embedding and GRU layers for both MCL and AC training phases. We train all models with Adam using the learning rate of 0.001. We stop training when there is no improvement in the validation accuracy for 5 consecutive epochs (i.e., early stopping with patience=5). The batch size is set to 64. 
During testing, we evaluate the final performance using MCL only. AC is not involved in testing (except the mask $\bm{a}_t$ generated with the task id $t$ during training). For the baselines SRK, HAT, OWM, and EWC, we use the code provided by their authors (customized for text if needed) and adopt their original parameters.

\begin{table}[t]
\centering
\resizebox{0.6\columnwidth}{!}{%
\begin{tabular}{ccccc}
\specialrule{.2em}{.1em}{.1em} 
Models & All Tasks & Last Tasks & P-value & \#Paramters \\ 
\specialrule{.1em}{.05em}{.05em} 
\specialrule{.1em}{.05em}{.05em} 
ONE & 0.7846 & 0.7809 & 1.025e-7 & 25.2M \\
LSC & 0.8219 & 0.8246 & 1.581e-2 & --- \\
\specialrule{.1em}{.05em}{.05em} 

N-CL & 0.8339 & 0.8477 & 1.792e-3 & 25.2M \\
EWC & 0.6899 & 0.7187 & 1.542e-9 & 42.5M \\ 
OWM & 0.6983 & 0.7337 & 3.219e-15 & 30.0M \\
HAT & 0.6456 & 0.6938 & 1.861e-14 & 42.7M \\
SRK & 0.8282 & 0.8500 & 7.793e-6 & 3.4M \\
\textbf{KAN} & \textbf{0.8524} & \textbf{0.8799} & --- & 42.4M \\ 
\specialrule{.1em}{.05em}{.05em} 
\end{tabular}%
}
\caption{Average accuracy of different models. \#Parameters refers to the number of parameters. LSC is based on Naive Bayes whose number of parameters is the sum of the number of unique words (vocabulary) in each dataset multiplied by the number of classes, which is 2 in our case. Paired $t$-test is used to test the significance of the result of KAN against that of each baseline for ALL Tasks. P-values are given {\color{black}in column 4.}} 
\label{tab:OverallAccuracy}
\end{table}

\subsection{Results}
\label{sec:results}


\vspace{+4mm}
\noindent
\textbf{Average Results.} 
We first report the average results of all compared models to show the effectiveness of the proposed KAN. 
Since the order of the tasks may have an impact on the final results of CL, we randomly choose and run 10 sequences and average their results. We also ensured that the last tasks in the 10 sequences are different. 
Table~\ref{tab:OverallAccuracy} gives the average accuracy of all systems. {\color{black}Column 2 gives the average result over 24 tasks for each model after all tasks are learned. Column 3 gives the accuracy result of the last task for each model. Note here that the Last Task results and the All Tasks results are not comparable because each Last Task result is the average of the 10 last tasks in the 10 random task sequences, while each ALL Tasks result is the average of all 24 tasks. Column 4 gives the p-value of the significance test to show that KAN outperforms all baselines (more discussion later). Column 5 gives the number of parameters of each model. Note that ONE and LSC are not continual learning (CL) systems and have no CF issue. The rest are continual learning systems. 

We first observe that KAN's result is markedly better than that of every baseline. Since the traditional lifelong learning method LSC mainly aims to improve the last task's performance, for the All Tasks column, we give its best result, i.e., putting each of the 24 tasks as the last task. Even under this favorable setting, its result is considerably poorer than that of KAN. 

It is interesting to know that comparing to ONE, naive CL (N-CL) does not show accuracy degradation on average even without a mechanism to deal with CF (forgetting). N-CL is actually significantly better than ONE (they use exactly the same network). As mentioned earlier, this is because sentiment analysis tasks are similar to each other and can mostly help one another. That is, CF is not a major issue in CL for sentiment analysis. In fact, N-CL also outperforms SRK on ALL Tasks. This can be explained by the fact that SRK does not adapt the past knowledge or allow backward transfer as discussed in Section 2. SRK's main goal was to improve the last task's performance rather than those of all tasks. 
  
Regarding the continual learning (CL) baselines, EWC, OWM and HAT, they perform poorly and are even worse than ONE, which is not surprising because
their networks are primarily designed to preserve knowledge learned for each of the previous tasks. This makes it hard for them to exploit knowledge sharing to improve all tasks. 


}

\textbf{Significance Test.} To show that our results from KAN are significantly better than those of baselines, we conduct a paired t-test. We test KAN against each of the baselines based on the results of All Tasks from the 10 random sequences. All p-values are far below {\color{black}0.05} (see Table \ref{tab:OverallAccuracy}), which indicates that KAN is significantly better than every baseline. 

Table~\ref{tab:OverallAccuracy} also includes the number of parameters of each neural model (Column 4). A large fraction of the parameters for KAN is due to the mask for each task. Note that the similar numbers of parameters of models by no means indicate the models are similar (see Section 2). 

\textbf{Ablation Study.} KAN has two components, AC and MCL. To evaluate the effectiveness of AC, we can remove AC to test KAN. However, the binary mask in KAN needs AC to train. Without AC, it will have no mask and KAN is the same {\color{black}as} N-CL. KAN is significantly better than N-CL as shown in Table~\ref{tab:OverallAccuracy}. MCL is the main module that performs continual learning and cannot be removed.

\begin{table*}[t]
\centering
\resizebox{\textwidth}{!}{%
\begin{tabular}{cccccccc}
\specialrule{.2em}{.1em}{.1em} 
Task (product category) & SRK & HAT & OWM & EWC & N-CL & ONE & \textbf{KAN} \\ 
\specialrule{.1em}{.05em}{.05em} 
\specialrule{.1em}{.05em}{.05em} 
Amazon\_Instant\_Video & 0.7776 & 0.6297 & 0.6792 & 0.6589 & 0.7989 & 0.7859 & \textbf{0.8293} \\ 
Apps\_for\_Android & 0.8236 & 0.6351 & 0.7044 & 0.6634 & 0.8431 & 0.8101 & \textbf{0.8531} \\ 
Automotive & 0.8049 & 0.6728 & 0.6849 & 0.6874 & 0.8086 & 0.6917 & \textbf{0.8335} \\ 
\specialrule{.1em}{.05em}{.05em} 

Baby & 0.8617 & 0.6947 & 0.7175 & 0.6997 & 0.8703 & 0.8020 & \textbf{0.8870} \\ 
Beauty & 0.8758 & 0.6684 & 0.7333 & 0.6816 & 0.8752 & 0.8081 & \textbf{0.8912} \\ 
Books & \textbf{0.8517} & 0.6260 & 0.7078 & 0.7112 & 0.8165 & 0.8101 & 0.8337 \\ 
\specialrule{.1em}{.05em}{.05em} 

CDs\_and\_Vinyl & 0.7740 & 0.5666 & 0.6487 & 0.6412 & 0.7937 & 0.7152 & \textbf{0.7970} \\ 
Cell\_Phones\_and\_Accessories & 0.8277 & 0.6363 & 0.7163 & 0.6844 & 0.8489 & 0.7899 & \textbf{0.8679} \\ 
Clothing\_Shoes\_and\_Jewelry & 0.8520 & 0.6678 & 0.7221 & 0.7124 & 0.8701 & 0.8727 & \textbf{0.8814} \\ 
\specialrule{.1em}{.05em}{.05em} 
Digital\_Music & 0.7480 & 0.5656 & 0.6335 & 0.6159 & \textbf{0.7717} & 0.7232 & 0.7706 \\ 
Electronics & \textbf{0.8457} & 0.6472 & 0.6895 & 0.6546 & 0.8242 & 0.7636 & 0.8359 \\ 
Grocery\_and\_Gourmet\_Food & 0.8800 & 0.6664 & 0.7203 & 0.7354 & 0.8686 & 0.8242 & \textbf{0.8828} \\ 
\specialrule{.1em}{.05em}{.05em} 
Health\_and\_Personal\_Care & 0.8076 & 0.6095 & 0.6642 & 0.6605 & 0.8235 & 0.7131 & \textbf{0.8335} \\ 
Home\_and\_Kitchen & 0.8577 & 0.6920 & 0.7289 & 0.7407 & 0.8595 & 0.8081 & \textbf{0.8812} \\ 
Kindle\_Store & 0.8500 & 0.6427 & 0.7060 & 0.6966 & 0.8324 & 0.8505 & \textbf{0.8584} \\ 
\specialrule{.1em}{.05em}{.05em} 
Movies\_and\_TV & 0.8377 & 0.6317 & 0.7053 & 0.6986 & 0.8278 & 0.7879 & \textbf{0.8527} \\ 
Musical\_Instruments & 0.8142 & 0.7595 & 0.7456 & 0.7570 & 0.8677 & 0.8351 & \textbf{0.8851} \\
Office\_Products & 0.8180 & 0.6195 & 0.6664 & 0.6592 & 0.8142 & 0.7374 & \textbf{0.8346} \\ 
\specialrule{.1em}{.05em}{.05em} 
Patio\_Lawn\_and\_Garden & 0.8130 & 0.6406 & 0.6543 & 0.7033 & 0.8308 & 0.7833 & \textbf{0.8363} \\ 
Pet\_Supplies & 0.7936 & 0.6289 & 0.6840 & 0.6720 & 0.8255 & 0.7556 & \textbf{0.8481} \\ 
Sports\_and\_Outdoors & 0.8420 & 0.6631 & 0.7045 & 0.6953 & 0.8384 & 0.7939 & \textbf{0.8696} \\ 
\specialrule{.1em}{.05em}{.05em} 
Tools\_and\_Home\_Improvement & 0.8300 & 0.6530 & 0.6904 & 0.6832 & 0.8408 & 0.7515 & \textbf{0.8640} \\ 
Toys\_and\_Games & 0.8500 & 0.6700 & 0.7368 & 0.7379 & 0.8644 & 0.8202 & \textbf{0.8744} \\ 
Video\_Games & 0.8397 & 0.6502 & 0.7147 & 0.7067 & 0.8381 & 0.7980 & \textbf{0.8557} \\ 
\specialrule{.1em}{.05em}{.05em} 
\end{tabular}%
}
\caption{Individual task accuracy of each model, after having trained on all tasks. The number in bold in each row is the best accuracy of the row. }
\label{tab:IndividualAccuracy}
\end{table*}

\subsubsection{Individual Task Results.} 
To gain more insights, we report the individual task results in Table~\ref{tab:IndividualAccuracy} for all continual learning baselines and KAN. We also include ONE for comparison. 
Note that each task result for a model is the average of the results from the 10 random sequences (except ONE). 

Table 2 shows that KAN gives the best accuracy for 21 out of 24 tasks. In those tasks where KAN does not give the best, KAN's performances are competitive.  Hence, we can conclude that KAN is both highly accurate and robust. 

Regarding SRK, it performs the best in only $2$ tasks. However, these two tasks' results are only slightly better than those of KAN. These clearly indicate SRK is weaker than KAN. For the other continual learning baselines: HAT, OWM and EWC, their performances are consistently worse even than ONE. This is expected because their goal is to protect each of the ONE's results and such protections are not perfect and thus can still result in some CF (forgetting). 

\begin{table*}[t!]
\centering
\resizebox{0.7\textwidth}{!}{%
\begin{tabular}{cccccc}
\specialrule{.2em}{.1em}{.1em} 
\multirow{2}{*}{Tasks} & \multirow{2}{*}{ONE} & \multicolumn{2}{c}{N-CL} & \multicolumn{2}{c}{\textbf{KAN}} \\ 
 &  & Forward & Backward & \textbf{Forward} & \textbf{Backward} \\
\specialrule{.1em}{.05em}{.05em} 
\specialrule{.1em}{.05em}{.05em} 
First 6 tasks & 0.7846 & 0.7937 & 0.7990 & \textbf{0.8068} & \textbf{0.8132} \\ 
First 12 tasks & 0.7865 & 0.8135 & 0.8199 & \textbf{0.8314} & \textbf{0.8390} \\ 
First 18 tasks & 0.7870 & 0.8253 & 0.8327 & \textbf{0.8424} & \textbf{0.8501} \\ 
First 24 tasks & 0.7846 & 0.8302 & 0.8339 & \textbf{0.8471} & \textbf{0.8524} \\ 
\specialrule{.1em}{.05em}{.05em} 
\end{tabular}%
}
\caption{Effects of forward and backward knowledge transfer of KAN.
We give progressive results after 6, 12, 18, and 24 tasks have been learned respectively.
}
\label{tab:KnowledgeTransfer}
\end{table*}

\subsubsection{Effectiveness of Forward and Backward Knowledge Transfer.} 
{\color{black}From Tables \ref{tab:OverallAccuracy} and \ref{tab:IndividualAccuracy}, we can already see that KAN is able to exploit shared knowledge to improve learning of similar tasks. Here, we want to show whether the forward knowledge transfer and the backward knowledge transfer are indeed effective. 

Table~\ref{tab:KnowledgeTransfer} shows the accuracy results progressively after every 6 tasks have been learned. In the second row, we give the results after 6 tasks have been learned sequentially. Each accuracy result in the Forward column is the overall average of the 6 tasks when each of them was first learned in each of the 10 random runs, which indicate the forward transfer effectiveness because from the second task, the system can already start to leverage the previous knowledge to help learn the new task. By comparing with the corresponding results of ONE {\color{black}and N-CL}, we can see forward transfer of KAN is indeed effective. N-CL also has the positive forward transfer effect, but KAN does better. Each result in the Backward column shows the average test accuracy after all 6 tasks have been learned (over 10 random runs). By comparing with the corresponding result in the Forward column, we can see that backward transfer of KAN is also effective, which means that learning of later tasks can help improve the earlier tasks automatically. The same is also true for N-CL, although KAN does better. Rows 3, 4, and 5 show the corresponding results after 12, 18, and 24 tasks have been learned, respectively. 

We also observe that forward transfer is much more effective than the backward transfer. This is expected because forward transfer leverages the previous knowledge first and backward transfer can improve only after the forward transfer has made significant improvements. Furthermore, backward transfer also has the risk of causing some forgetting for the previous tasks because the previous task data are no longer available to prevent it. }




\section{Conclusion and Future Work}
\label{sec:conclusion}
This paper proposed KAN, a novel neural network for continual learning (CL) of a sequence of sentiment classification (SC) tasks. Previous CL models primarily focused on dealing with catastrophic forgetting (CF). As we have seen in the experiment section, CF is not a major issue for continual sentiment classification because the SC tasks are similar to each other and have a significant amount of shared knowledge among them. KAN thus focuses on improving the learning accuracy by exploiting the shared knowledge via forward and backward knowledge transfer. KAN achieves these goals using a knowledge base and a knowledge accessibility network. 
The effectiveness of KAN was demonstrated by empirically comparing it with state-of-the-art CL approaches. KAN's bi-directional knowledge transfer for CL significantly improves its results for SC. Our future work will improve its accuracy and adapt it for other types of data.

\section*{Acknowledgments}
This work was supported in part by two grants from National Science Foundation: IIS-1910424 and IIS-1838770, and a research gift from Northrop Grumman.

\bibliography{anthology,acl2020}

\begin{thebibliography}{10}
\providecommand{\url}[1]{\texttt{#1}}
\providecommand{\urlprefix}{URL }
\providecommand{\doi}[1]{https://doi.org/#1}

\bibitem{aharoni2017}
Aharoni, R., Goldberg, Y.: Morphological inflection generation with hard
  monotonic attention. In: ACL. pp. 2004--2015. Association for Computational
  Linguistics, Vancouver, Canada (2017)

\bibitem{Chaudhry2019ICLR}
Chaudhry, A., Ranzato, M., Rohrbach, M., Elhoseiny, M.: Efficient lifelong
  learning with {A-GEM}. In: ICLR (2019)

\bibitem{chen2014topic}
Chen, Z., Liu, B.: Topic modeling using topics from many domains, lifelong
  learning and big data. In: ICML. pp. 507--515 (2014)

\bibitem{chen2018lifelong}
Chen, Z., Liu, B.: Lifelong machine learning. Synthesis Lectures on Artificial
  Intelligence and Machine Learning  \textbf{12}(3),  1--207 (2018)

\bibitem{DBLP:conf/acl/ChenM015}
Chen, Z., Ma, N., Liu, B.: Lifelong learning for sentiment classification. In:
  ACL. pp. 750--756 (2015)

\bibitem{fernando2017pathnet}
Fernando, C., Banarse, D., Blundell, C., Zwols, Y., Ha, D., Rusu, A.A.,
  Pritzel, A., Wierstra, D.: Pathnet: Evolution channels gradient descent in
  super neural networks. CoRR  \textbf{abs/1701.08734} (2017)

\bibitem{He2018overcoming}
He, X., Jaeger, H.: Overcoming catastrophic interference using conceptor-aided
  backpropagation. In: ICLR (2018)

\bibitem{Kamra2017deep}
Kamra, N., Gupta, U., Liu, Y.: Deep generative dual memory network for
  continual learning. CoRR  \textbf{abs/1710.10368} (2017)

\bibitem{Kirkpatrick2017overcoming}
Kirkpatrick, J., Pascanu, R., Rabinowitz, N.C., Veness, J., Desjardins, G.,
  Rusu, A.A., Milan, K., Quan, J., Ramalho, T., Grabska{-}Barwinska, A.,
  Hassabis, D., Clopath, C., Kumaran, D., Hadsell, R.: Overcoming catastrophic
  forgetting in neural networks. CoRR  \textbf{abs/1612.00796} (2016)

\bibitem{kornell2009unsuccessful}
Kornell, N., Hays, M.J., Bjork, R.A.: Unsuccessful retrieval attempts enhance
  subsequent learning. Journal of Experimental Psychology: Learning, Memory,
  and Cognition  \textbf{35}(4), ~989 (2009)

\bibitem{DBLP:journals/corr/abs-1909-08383}
Lange, M.D., Aljundi, R., Masana, M., Parisot, S., Jia, X., Leonardis, A.,
  Slabaugh, G.G., Tuytelaars, T.: Continual learning: {A} comparative study on
  how to defy forgetting in classification tasks. CoRR  \textbf{abs/1909.08383}
  (2019)

\bibitem{DBLP:conf/nips/LeeKJHZ17}
Lee, S., Kim, J., Jun, J., Ha, J., Zhang, B.: Overcoming catastrophic
  forgetting by incremental moment matching. In: NeurIPS. pp. 4652--4662 (2017)

\bibitem{liu2015sentiment}
Liu, B.: Sentiment analysis: Mining opinions, sentiments, and emotions.
  Cambridge University Press (2015)

\bibitem{Lopez2017gradient}
Lopez{-}Paz, D., Ranzato, M.: Gradient episodic memory for continual learning.
  In: NeurIPS. pp. 6467--6476 (2017)

\bibitem{DBLP:conf/dasfaa/LvWLCZ19}
Lv, G., Wang, S., Liu, B., Chen, E., Zhang, K.: Sentiment classification by
  leveraging the shared knowledge from a sequence of domains. In: DASFAA. pp.
  795--811 (2019)

\bibitem{DBLP:conf/cvpr/MallyaL18}
Mallya, A., Lazebnik, S.: Packnet: Adding multiple tasks to a single network by
  iterative pruning. In: CVPR. pp. 7765--7773 (2018)

\bibitem{mccloskey1989catastrophic}
McCloskey, M., Cohen, N.J.: Catastrophic interference in connectionist
  networks: The sequential learning problem. In: Psychology of learning and
  motivation, vol.~24, pp. 109--165. Elsevier (1989)

\bibitem{Parisi2019continual}
Parisi, G.I., Kemker, R., Part, J.L., Kanan, C., Wermter, S.: Continual
  lifelong learning with neural networks: {A} review. Neural Networks
  \textbf{113},  54--71 (2019)

\bibitem{pontiki2016semeval}
Pontiki, M., Galanis, D., Papageorgiou, H., Androutsopoulos, I., Manandhar, S.,
  Al-Smadi, M., Al-Ayyoub, M., Zhao, Y., Qin, B., De~Clercq, O., et~al.:
  Semeval-2016 task 5: Aspect based sentiment analysis. In: SemEval (2016)

\bibitem{Rebuffi2017}
Rebuffi, S., Kolesnikov, A., Sperl, G., Lampert, C.H.: icarl: Incremental
  classifier and representation learning. In: CVPR. pp. 5533--5542 (2017)

\bibitem{Rostami2019ijcai}
Rostami, M., Kolouri, S., Pilly, P.K.: Complementary learning for overcoming
  catastrophic forgetting using experience replay. In: IJCAI. pp. 3339--3345
  (2019)

\bibitem{ruvolo2013ella}
Ruvolo, P., Eaton, E.: {ELLA:} an efficient lifelong learning algorithm. In:
  ICML. pp. 507--515 (2013)

\bibitem{Seff2017continual}
Seff, A., Beatson, A., Suo, D., Liu, H.: Continual learning in generative
  adversarial nets. CoRR  \textbf{abs/1705.08395} (2017)

\bibitem{DBLP:conf/icml/SerraSMK18}
Serr{\`{a}}, J., Suris, D., Miron, M., Karatzoglou, A.: Overcoming catastrophic
  forgetting with hard attention to the task. In: ICML. pp. 4555--4564 (2018)

\bibitem{Shin2017continual}
Shin, H., Lee, J.K., Kim, J., Kim, J.: Continual learning with deep generative
  replay. In: NeurIPS. pp. 2990--2999 (2017)

\bibitem{ShuXuLiu2017}
Shu, L., Xu, H., Liu, B.: Lifelong learning {CRF} for supervised aspect
  extraction. In: ACL. pp. 148--154 (2017)

\bibitem{Silver2013}
Silver, D.L., Yang, Q., Li, L.: Lifelong machine learning systems: Beyond
  learning algorithms. In: AAAI (2013)

\bibitem{hao2019forward}
Wang, H., Liu, B., Wang, S., Ma, N., Yang, Y.: Forward and backward knowledge
  transfer for sentiment classification. In: ACML. pp. 457--472 (2019)

\bibitem{shuai2018lifelong}
Wang, S., Lv, G., Mazumder, S., Fei, G., Liu, B.: Lifelong learning memory
  networks for aspect sentiment classification. In: {IEEE} International
  Conference on Big Data, Big Data. pp. 861--870 (2018)

\bibitem{xia2017distantly}
Xia, R., Jiang, J., He, H.: Distantly supervised lifelong learning for
  large-scale social media sentiment analysis. {IEEE} Trans. Affective
  Computing  \textbf{8}(4),  480--491 (2017)

\bibitem{DBLP:conf/icml/XuBKCCSZB15}
Xu, K., Ba, J., Kiros, R., Cho, K., Courville, A.C., Salakhutdinov, R., Zemel,
  R.S., Bengio, Y.: Show, attend and tell: Neural image caption generation with
  visual attention. In: ICML. pp. 2048--2057 (2015)

\bibitem{zeng2019continuous}
Zeng, G., Chen, Y., Cui, B., Yu, S.: Continuous learning of context-dependent
  processing in neural networks. Nature Machine Intelligence  (2019)

\end{thebibliography}
\bibliographystyle{splncs04}

\end{document}